\documentclass[runningheads,a4paper]{llncs}

\usepackage{graphicx}
\usepackage{booktabs}
\usepackage{multirow}
\usepackage{amsmath,amssymb}
\usepackage{url}
\usepackage{algorithm}
\usepackage{algorithmic}
\usepackage{subcaption}
\usepackage[hidelinks]{hyperref}

\newcommand{\argmax}{\operatorname*{arg\,max}}

\newcommand{\Man}{d_{\text{Man}}}

\usepackage{soul,todonotes}
\usepackage{orcidlink}
\renewcommand{\orcidID}[1]{\orcidlink{#1}}

\begin{document}

\title{Reading Order Inference\\ for Complex Document Layouts}
\titlerunning{Reading Order Inference}

\author{Iddo Hakim\orcidID{0009-0000-3860-8941} \and
Sharva Gogawale\orcidID{0009-0000-5230-5197} \and
Omer Ventura \and
Gal Grudka \and
Daria Vasyutinsky-Shapira\orcidID{0000-0002-4257-7882} \and
Berat Kurar-Barakat\orcidID{0000-0002-7240-7286} \and
Nachum Dershowitz}
\authorrunning{I. Hakim et al.}
\institute{School of Computer Science and AI, Tel Aviv University, Ramat Aviv, Israel\\
\email{\{iddoh,sharvag,omerventura,galgrudka\}@mail.tau.ac.il}\\
\email{\{dariashap,berat,nachumd\}@tauex.tau.ac.il}}
\pagestyle{plain} 
\sloppy
\maketitle
\begin{abstract}
Reading order inference remains a critical bottleneck in the digitization of complex historical
manuscripts, where pages contain multiple spatially interleaved reading streams, the canonical
example being the \emph{Glossa Ordinaria} layout, in which a central text is surrounded by
commentaries that wrap around it in non-rectangular, non-convex regions. We present a
\emph{training-free}, graph-based framework: each OCR text line becomes a node in a directed
candidate-transition graph, edges are scored by a weighted additive ensemble of two lightweight
language-model signals (causal language model conditional likelihood and BERT next-sentence
prediction, NSP; a third sentence-embedding signal was evaluated but did not improve reading order),
and the global reading order is recovered as a degree-constrained directed path cover. To avoid the cascading ``edge-theft'' failures of greedy
edge selection, we propose a max-regret inference rule that prioritizes commitments with high
opportunity cost. We evaluate on synthetic Glossa Ordinaria grid layouts, on \emph{23 ALTO page geometries}
(10 historical source pages plus mirrored and flipped variants), and on a \emph{140-page multi-column
English subset of OmniDocBench}, comparing our method against the canonical recursive XY-cut
(PaddleOCR PP-StructureV3) and two LayoutReader variants (layout-only and text+layout) on
identical inputs. On wrap-around Glossa layouts our method recovers $95\%$ of ground-truth
successor edges on average vs.\ XY-cut's $50\%$; on the OmniDocBench multi-column subset it
reaches $88\%$ macro edge accuracy versus XY-cut's $75\%$ and LayoutReader's $25\%$. The
LayoutReader baselines transfer poorly due to a word-level vs.\ line-level granularity mismatch.
We additionally verify mirror-invariance under horizontal and vertical page reflections: Our
method changes by less than $1$ percentage point, classical XY-cut by $2$ points,
and LayoutReader-T by up to $8$ points.
\end{abstract}

\keywords{Reading order inference \and Document layout analysis \and Non-Manhattan layouts \and
Training-free \and Max-regret inference \and Historical manuscripts \and XY-cut \and LayoutReader \and
Mirror invariance}

\begin{figure}[t]
  \centering
  \includegraphics[width=0.166\linewidth]{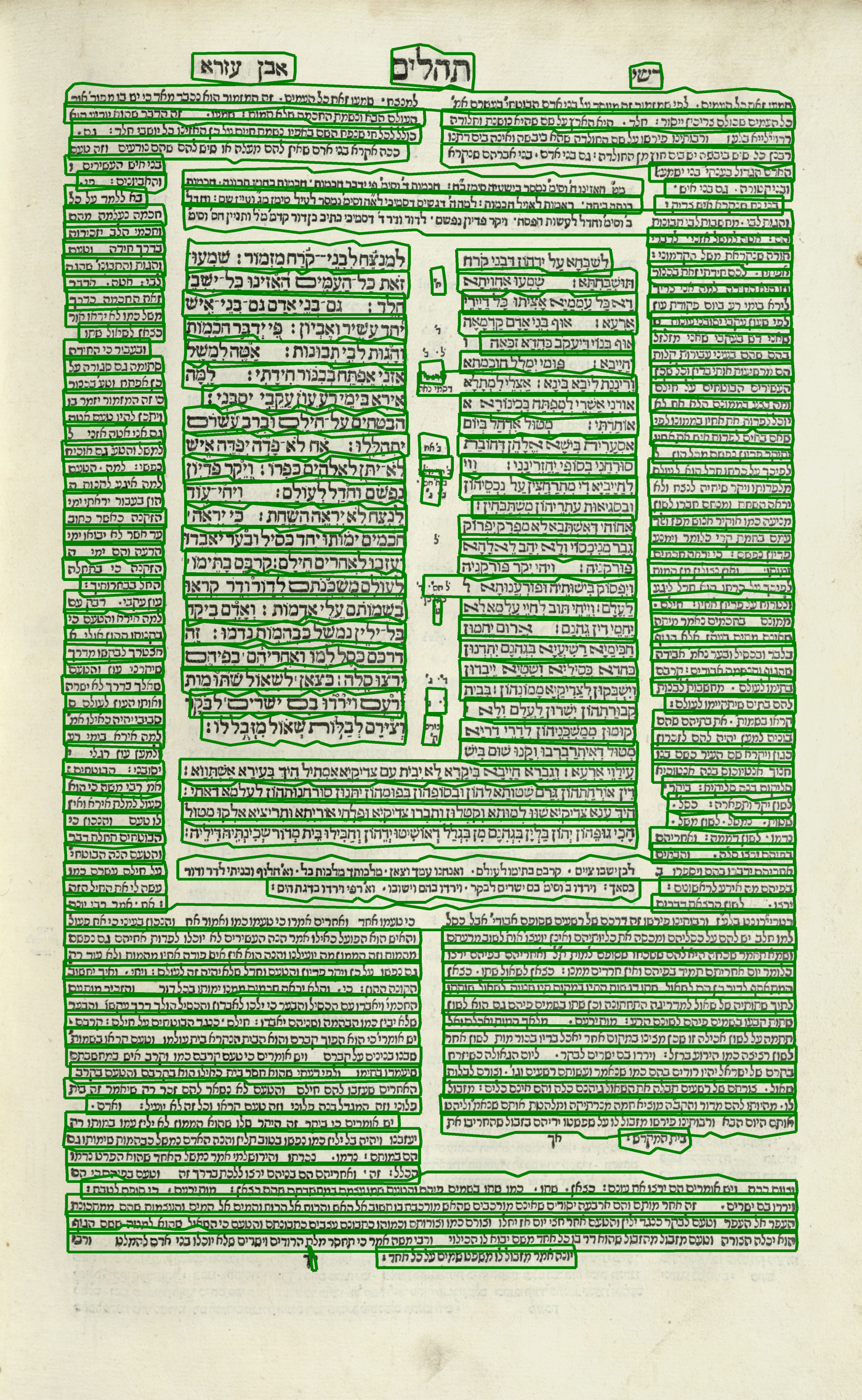}\hspace{1em}%
  \includegraphics[width=0.2\linewidth]{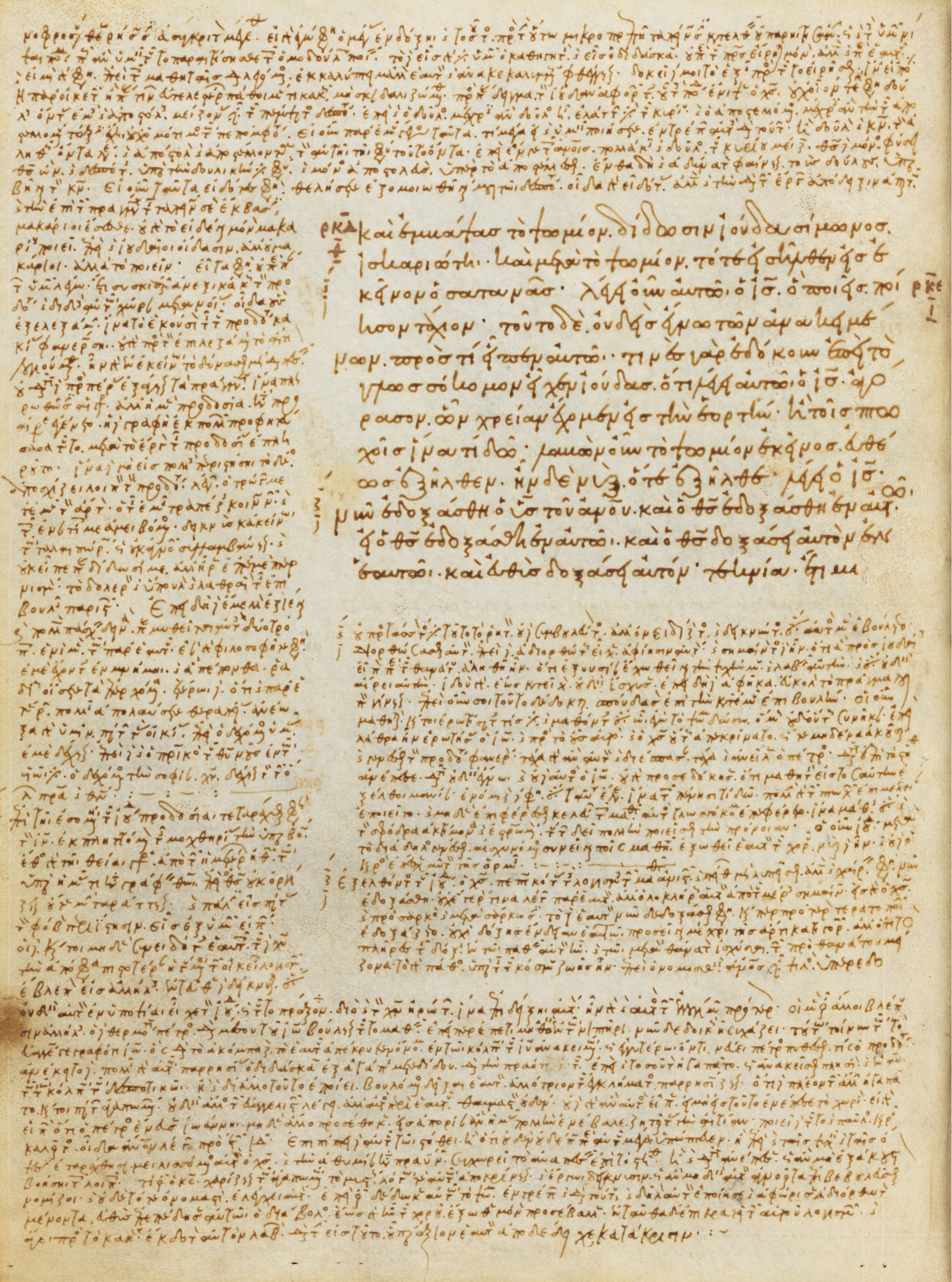}
  \caption{Two examples of non-Manhattan layouts. \textit{Left:} a printed Hebrew Bible page, main text
  flanked by an Aramaic translation, two commentaries wrapping around them, Masorah parva as abbreviated
  notes in an internal margin, and Masorah magna spanning the top and bottom margins, all automatically
  (and imperfectly) line-segmented. \textit{Right:} a page of a manuscript (Codex Bodmer 25) of the Greek
  Bible with two regions of catena commentary.}
  \label{fig:talmud_scan_boxes}
\end{figure}

\section{Introduction}
Reading order inference reconstructs the intended sequential flow of text elements detected by OCR.
Accurate order is a prerequisite for searchable, continuous text streams and downstream document
understanding; yet, it remains fragile whenever a page contains multiple spatially interleaved
narratives.

For simple single-column pages, ordering is recovered by a top-to-bottom scan. Visually complex
documents often contain \emph{multiple partially independent reading streams}: parallel columns,
marginalia, interlinear glosses, side notes, and figure captions, where \emph{geometry alone is
underdetermined}: several plausible successors may be nearby in space but belong to different
narrative tracks.

Historical manuscripts amplify these difficulties. A canonical extreme is the \emph{Glossa Ordinaria}
page, where a central text is surrounded by multiple commentaries that wrap around it in
non-rectangular, sometimes non-convex regions. Lines from different streams may be spatially
interleaved, and human readers rely primarily on \emph{semantic continuity} to remain within a stream.
Classical geometric heuristics, including recursive XY-cut and proximity-based clustering, implicitly
assume that spatial adjacency implies sequential adjacency, which often fails on such multi-text
layouts \cite{nagy1984icpr,ogorman1993docstrum,meunier2005xycut}. We treat the canonical recursive
XY-cut as our primary geometric baseline.

A natural alternative is to learn reading order from annotated documents, and modern multimodal models
can be effective in-domain. But reading-order supervision is typically collected at the word/token
level (e.g.\ ReadingBank for LayoutReader~\cite{wang2021layoutreader}), which differs from the OCR
text-line atomic unit produced by historical OCR engines. As Section~\ref{sec:experiments} shows,
off-the-shelf LayoutReader variants, both layout-only and text-plus-layout (LayoutReader-T), transfer
poorly to line-level historical inputs without retraining, even when fed the exact same boxes and text
as our method. High-quality reading-order supervision for rare historical layouts is also expensive to
obtain. This motivates \emph{training-free} methods that exploit general linguistic priors already
present in pretrained language models.

\paragraph{Goal.} Recover reading order in complex, non-Manhattan, multi-stream layouts without
annotated reading-order datasets or layout-specific fine-tuning.

\paragraph{Key idea.} Pretrained language models, even small ones, encode strong local coherence
signals. We model a page as a directed graph of candidate ``next-line'' transitions; nodes are OCR
lines and edge weights quantify semantic continuity. Given this scored graph, we search for a globally
consistent set of successor relations under the constraint that each line has at most one predecessor
and one successor. Throughout, we isolate layout inference from OCR noise by evaluating on known line
boxes with controlled text; end-to-end robustness to OCR errors is out of scope
(Section~\ref{sec:limitations}).

\subsection*{Contributions}
\begin{enumerate}
  \item \textbf{Graph formulation.} Reading order as a degree-constrained directed path cover over OCR
  lines, enabling multiple disjoint reading streams under one-predecessor / one-successor constraints.
  \item \textbf{Training-free semantic scoring.} A weighted additive ensemble of causal LM conditional
  likelihood and BERT next-sentence prediction. We additionally evaluate sentence-embedding cosine
  similarity and find it does not improve reading order (Section~\ref{sec:Results}).
  \item \textbf{Regret-based inference.} A max-regret edge-selection algorithm that reduces greedy
  myopia by prioritizing decisions with high opportunity cost.
  \item \textbf{Expanded evaluation across synthetic, historical, and public-benchmark layouts.}
  Synthetic ``Glossa Ordinaria layout'' pages~\cite{Talmudic} created by interlacing distinct Project
  Gutenberg books%
  \footnote{\url{https://www.gutenberg.org}} into non-convex topologies; \emph{23 ALTO page geometries}~\cite{alto_loc} from 10 historical
  source pages and their mirrored and flipped variants, a substantial expansion over prior
  small-scale studies; and a 140-page English multi-column subset of
  OmniDocBench~\cite{omnidocbench} for direct comparison on a public benchmark.
  \item \textbf{Multi-baseline comparison.} We benchmark against the canonical recursive
  XY-cut~\cite{nagy1984icpr} (PaddleOCR PP-StructureV3), LayoutReader~\cite{wang2021layoutreader}
  (layout-only), and LayoutReader-T (text+layout), on identical OCR-line inputs, and verify
  mirror-invariance under horizontal and vertical page reflections.
\end{enumerate}
\section{Related Work}
Reading order inference sits at the intersection of page layout analysis and document structure
recovery. While many documents admit a near-linear scan, visually complex pages may contain multiple
interleaved reading streams (marginalia, parallel commentaries) for which geometric adjacency is an
unreliable proxy for sequential adjacency.

\subsection{Geometric Heuristics}
Early reading-order approaches rely on spatial regularities: projection-profile and recursive
partitioning (XY-cut)~\cite{nagy1984icpr}, nearest-neighbor clustering
(Docstrum)~\cite{ogorman1993docstrum}, and efficiency-tuned XY-cut variants~\cite{meunier2005xycut}.
These work well on Manhattan layouts but degrade when streams are spatially interleaved and region
boundaries are non-convex, as in many historical manuscripts. We use the canonical recursive XY-cut as
implemented in PaddleOCR's PP-StructureV3 pipeline as our geometric baseline
(Section~\ref{sec:experiments}), so comparisons reflect a standard, citable implementation rather than
a re-implementation.

\subsection{Learning-Based and Semantic Approaches}
Recent methods learn reading order as sequence generation or as pairwise relation prediction over
layout entities, with multimodal inputs and supervised datasets
\cite{xu2020layoutlm,wang2021layoutreader,wang2023sparsegraph}.
LayoutReader~\cite{wang2021layoutreader} is trained on ReadingBank, a corpus of \emph{word-level}
layout and reading order from rendered DOCX files; we evaluate both its layout-only and
text-plus-layout (LayoutReader-T) configurations on our OCR-line and OmniDocBench paragraph inputs
(Section~\ref{sec:experiments}), where the word-to-line and word-to-paragraph granularity mismatch
becomes a concrete limitation. End-to-end multimodal parsers such as Dots.OCR and PaddleOCR-VL
operate at line granularity but bundle detection, recognition, and reading order into a single
pipeline and do not expose their reading-order module for evaluation on externally supplied boxes,
so a clean apples-to-apples comparison is not currently possible. Such systems are typically
accurate in-distribution but depend on labeled reading order and may require fine-tuning for rare
historical layouts. In NLP, sentence ordering and discourse coherence use local continuity cues and
global optimization~\cite{barzilay2008coherence,li2014discourse}, but document reading order differs
in three ways: the units are OCR lines (not sentences), the problem is constrained by 2D candidate
adjacency, and multi-stream outputs (sets of disjoint paths) are often required rather than a single
permutation. Our approach is training-free, uses pretrained language models as local semantic
oracles, and combines them with explicit graph inference under degree constraints.

\subsection{Reading-Order Representation and Evaluation}
Reading order may be non-linear or partially specified, and evaluation becomes non-trivial when
segmentation differs across systems~\cite{clausner2013readingorder}.
Public benchmarks such as OmniDocBench~\cite{omnidocbench} cover diverse modern documents (academic
papers, financial reports, multi-column layouts) and evaluate ordering over document components or
paragraph-like regions rather than OCR lines. For our wrap-around, non-Manhattan target setting,
region decomposition is itself unreliable: regions may merge or split incorrectly even when
individual line boxes remain usable (a Sayre's-paradox-like situation, where correct region ordering
needs correct regions, but finding them depends on the reading streams). We therefore evaluate
directly over OCR lines and successor relations on our 23-page ALTO corpus, and complement this with
an evaluation on the multi-column English subset of OmniDocBench at its native paragraph granularity
(Section~\ref{sec:omnidocbench}). The graph formulation operates on (bbox, text) tuples regardless of
unit size, so both granularities are handled with the same model. ALTO and similar archival formats
can encode layout structure and optionally include ordering, but reading order is frequently missing
or unreliable in complex manuscripts, motivating automatic inference.

\section{Problem Formulation}
We assume an OCR system produces a set of $N$ text lines.

\subsection{Candidate Transition Graph}
We build a directed candidate graph $G=(V,E_{\text{cand}})$ with $V=\{1,\dots,N\}$.
An edge $(u,v)\in E_{\text{cand}}$ denotes that line $v$ is a plausible successor of line $u$.
Candidate generation controls both runtime and ambiguity; we use permissive ``next-column'' candidate sets to
stress-test inter-column continuation decisions.

\begin{itemize}
  \item \textit{In artificial grid layouts},
  each node belongs to a discrete column. For each node $u$, we connect it to \emph{all} nodes
  in the next column to the right. This produces a dense candidate set in which geometry alone cannot determine
  which inter-column transition is the correct continuation.
  \item \textit{In realistic layouts}, where nodes are derived from OCR text bounding boxes, we mimic the same
  stress test using bounding boxes: for each line $u$, we connect it to \emph{all} lines below $u$
  (same-column successors) and to \emph{all} lines whose left edge lies to the right of $u$'s right boundary
  (next-column successors). This permissive set makes both same-column continuations and wrap-around
  ``next-column'' transitions candidates, so geometry alone cannot determine the correct successor in
  complex page geometries.
\end{itemize}

\subsection{Degree Constraints and Path Cover}
Reading order within each stream is a directed path: each line has at most one successor and at most one
predecessor. We seek a subset $E \subseteq E_{\text{cand}}$ that maximizes the total score:
\begin{equation}
\label{eq:objective}
\max_{E \subseteq E_{\text{cand}}}
\sum_{(u,v)\in E} S(u,v)
\quad \text{s.t.}\quad
\deg^+(u)\le 1,\;\deg^-(v)\le 1\;\forall u,v.
\end{equation}
where $S(u,v)$ is the semantic continuity score for the edge.
Under the degree constraints alone, a maximizing edge set decomposes into a collection of disjoint directed
paths and (in general) directed cycles.
Because cycles do not correspond to valid reading orders, our inference procedures explicitly avoid cycle creation and return an acyclic path cover, possibly with
multiple disjoint streams.
\section{Training-Free Semantic Edge Scoring}
For each candidate edge $(u,v)$ we compute a semantic continuity score $S(u,v)$ using pretrained
models \emph{without fine-tuning}. Scores are computed independently per edge and cached, decoupling
expensive model inference from the downstream search algorithm.

\subsection{Causal Language Model Conditional Likelihood}
Let a causal language model (CLM) define token log-likelihood $\log P_{\text{CLM}}(\cdot)$ under a
fixed tokenizer. We instantiate the CLM with EleutherAI/pythia-410M and compute all CLM scores on a
GPU in inference mode without fine-tuning. Given two line fragments $t_u$ and $t_v$, we score the
per-token continuation likelihood of $t_v$ conditioned on $t_u$:
\begin{equation*}
s_{\text{clm}}(u,v)
=
\frac{1}{|t_v|}
\log P_{\text{CLM}}(t_v \mid t_u),
\end{equation*}
where $|t_v|$ denotes the number of tokens in $t_v$.
In practice, we compute this by concatenating the token sequences $[t_u ; t_v]$,
running one forward pass, and summing log-probabilities only over the token positions belonging to
$t_v$ (i.e., masking out the prefix $t_u$) \cite{radford2019gpt2}.

\paragraph{Context Truncation.}
Because causal models have a finite context window, we truncate $t_u$ to the most recent $L$ tokens
(we use a fixed $L$ across all experiments) so that $[t_u;t_v]$ fits within the model context.

\paragraph{Optional Frequency Normalization.}
Rare-token artifacts can inflate $s_{\text{clm}}$ by over-rewarding unlikely but highly specific
continuations. We therefore optionally normalize by the unconditional likelihood of $t_v$:
\begin{equation}
\label{eq:freqnorm}
\widetilde{s}_{\text{clm}}(u,v)
=
s_{\text{clm}}(u,v)
-
\kappa \cdot
\frac{1}{|t_v|}
\log P_{\text{CLM}}(t_v),
\end{equation}
with $\kappa\in[0,1]$ controlling the correction strength. When $\kappa=0$, no normalization is
applied. We use a different symbol ($\kappa$) here than for the ensemble blending weights below
($\alpha,\beta,\gamma$, see Sec.~\ref{sec:Results}) to make the two roles distinct.

\subsection{Next Sentence Prediction}
BERT's next sentence prediction (NSP) head produces $p_{\text{nsp}}(u,v)\in(0,1)$ indicating whether
$t_v$ is a plausible continuation of $t_u$ \cite{devlin2019bert}. NSP is the binary
sentence-continuity objective on which BERT was pretrained, which makes it a standard, training-free
probe for precisely the adjacency relation a reading-order edge encodes.
We convert this to an additive score in log space:
\begin{equation*}
s_{\text{nsp}}(u,v) = \log \big(\max(\varepsilon, p_{\text{nsp}}(u,v))\big),
\end{equation*}
where $\varepsilon$ is a small constant to avoid $\log(0)$.

\subsection{Sentence Embedding Similarity}
We compute embeddings $h_u,h_v$ using a sentence-transformer encoder \cite{reimers2019sbert} and
define:
\begin{equation*}
s_{\text{sim}}(u,v)=\cos(h_u,h_v).
\end{equation*}
Since cosine similarity can be negative, we use it directly as a bounded additive signal (no
logarithm). We include this signal for completeness; the ablation in Section~\ref{sec:Results} shows
it does not improve reading order, so all reported results set $w_{\text{sim}}=0$.

\subsection{Weighted Additive Ensemble}
We combine all signals with a weighted additive ensemble:
\begin{equation*}
S(u,v)
=
w_{\text{clm}}\cdot \widetilde{s}_{\text{clm}}(u,v)
+
w_{\text{nsp}}\cdot s_{\text{nsp}}(u,v)
+
w_{\text{sim}}\cdot s_{\text{sim}}(u,v)
-
\lambda\cdot \Man(u,v),
\end{equation*}
where $w_{\text{clm}}, w_{\text{nsp}}, w_{\text{sim}} \ge 0$ and $\lambda\ge 0$.
We use Manhattan distance between bounding-box centers,
$\Man(u,v)=|c_x(u)-c_x(v)|+|c_y(u)-c_y(v)|$, as an optional regularizer that discourages long-distance
``teleportation'' edges in very dense candidate graphs.
Unless stated otherwise, we set $\lambda=0$ and rely on candidate gating for locality.

\paragraph{Fixed-Parameter Transfer.}
The ensemble weights are selected once by a sweep on the synthetic grid task
(Section~\ref{sec:Results}) and then \emph{transferred unchanged} to all ALTO and OmniDocBench
experiments. The sweep selects $w_{\text{sim}}=0$ (the embedding never helped and degraded results as its weight
grew, Section~\ref{sec:Results}), so the deployed scorer uses only $(w_{\text{clm}}, w_{\text{nsp}})$.
We do not re-tune per dataset or per document: the Section~\ref{sec:experiments} numbers are a
fixed-parameter transfer test, not in-sample fits.

\section{Inference Algorithms}
Given candidate edges $E_{\text{cand}}$ and scores $S(u,v)$, we infer a degree-constrained directed path cover:
each node has at most one successor and at most one predecessor, yielding multiple disjoint reading streams.
The maximum-weight degree-constrained acyclic path cover of Eq.~\eqref{eq:objective} is a combinatorial
problem we do not solve to optimality; the algorithms below are fast heuristics, and max-regret in
particular carries no optimality guarantee, only reordering which decisions are committed first.

\subsection{Greedy Selection Baselines}
We consider two greedy constructions that repeatedly select high-scoring edges subject to feasibility:
(i) \emph{local greedy}, which processes nodes in a fixed scan order and assigns each node its best feasible successor,
and (ii) \emph{global greedy}, which repeatedly commits the highest-scoring feasible edge overall.
While fast, greedy methods often fail by \emph{edge theft}: an early incorrect edge consumes a node's in-degree,
making the correct predecessor later unreachable and causing cascading errors.

\subsection{Max-regret Edge Selection}
To reduce greedy myopia, we prioritize decisions with high opportunity cost.
For each unresolved node $u$, let $v_1$ and $v_2$ denote the top two \emph{feasible} outgoing candidates by score,
where feasibility means: (a) $u$ has no assigned successor, (b) $v$ has no assigned predecessor, and
(c) adding $(u,v)$ would not create a directed cycle in the current partial solution.
We define regret as:
\begin{equation}
\label{eq:regret}
\mathrm{Regret}(u)=S(u,v_1)-S(u,v_2).
\end{equation}
If a node has fewer than two feasible candidates, we set $\mathrm{Regret}(u)=0$ so that low-ambiguity nodes
(e.g., boundary nodes with a single remaining option) do not dominate the schedule.

At each iteration, we select the node with maximum regret and commit its best feasible edge.
If the best edge would form a cycle, we fall back to the next-best feasible candidate for that node.

\begin{algorithm}[t]
\caption{Max-regret inference (cycle-avoiding path cover)}
\label{alg:maxregret}
\begin{algorithmic}[1]
\STATE Input: candidate edges $E_{\text{cand}}$, scores $S(u,v)$
\STATE Output: edge set $E$ (acyclic path cover)
\STATE $E \leftarrow \emptyset$; initialize all nodes as unassigned-in and unassigned-out
\WHILE{there exists an unresolved node with at least one feasible outgoing candidate}
  \FORALL{unresolved nodes $u$}
    \STATE Let $\mathcal{C}(u)=\{v : (u,v)\in E_{\text{cand}} \text{ and } (u,v)\text{ is feasible}\}$
    \STATE Sort $\mathcal{C}(u)$ in descending order of $S(u,v)$ to get $v_1, v_2, \dots$
    \STATE $\mathrm{Regret}(u) \leftarrow S(u,v_1)-S(u,v_2)$ \COMMENT{or $0$ if $|\mathcal{C}(u)|<2$}
  \ENDFOR
  \STATE $u^\star \leftarrow \argmax_u \mathrm{Regret}(u)$ \COMMENT{tie-break by $S(u,v_1)$}
  \STATE Commit the best feasible edge $(u^\star, v)$ by scanning $v_1, v_2, \dots$ until one does not create a cycle
  \STATE Add the committed edge to $E$ and mark $u^\star$ assigned-out and $v$ assigned-in
\ENDWHILE
\STATE \RETURN $E$
\end{algorithmic}
\end{algorithm}

\paragraph{Complexity.}
With $N$ nodes and $M$ candidate edges, each iteration selects one edge and updates feasibility.
A straightforward implementation runs in $O(M\log d)$ time if per-node candidate lists are pre-sorted
($d$ is average out-degree), with near-constant-time cycle checks using union-find-style successor tracing.

\section{Experimental Setup}
\label{sec:experiments}
To isolate layout inference from OCR recognition noise, we evaluate on pages where the set of line
bounding boxes is known and text content is controlled. We use three complementary settings: (i) synthetic
grid layouts for controlled ablations over topology and candidate density, (ii) authentic historical
page geometries extracted from ALTO XML~\cite{alto_loc} that preserve realistic, irregular line placements
and non-rectangular region boundaries, and (iii) a multi-column English subset of the OmniDocBench public
benchmark~\cite{omnidocbench} for an apples-to-apples comparison on a community-recognized dataset.

\subsection{Synthetic Grid Complex Layouts}
We instantiate layouts on an $R\times C$ grid partitioned into disjoint regions (text streams) with
non-convex wrap-around topologies (L-, H-, Y-, and O-shapes). Nodes in each region are populated with
consecutive lines from a distinct Project Gutenberg book, producing spatially interleaved but
semantically independent streams. Ground-truth reading order
inside each region is a fixed deterministic traversal (row-major or column-major, held constant per
experiment).

\subsection{Historical Geometries from ALTO XML}
We parse \texttt{TextLine} elements (and polygons when available) from ALTO XML. Our corpus comprises
\emph{23 page geometries}: 10 historical source pages, 2 horizontal
mirrors, 10 vertical (top-bottom) flips, and one two-column ALTO as a Manhattan sanity check. The
mirror and flip variants are produced by reflecting bounding-box coordinates and re-deriving each
group's reading order under a column-major edge-aligned clustering rule, preserving the \emph{semantic}
reading order of the source page while breaking directional bias; they are used in
Section~\ref{sec:mirror_inv} as a mirror-invariance test.

Each node is populated with consecutive Project Gutenberg words up to the node's box width; the text
then continues into the next node in the region. RTL source layouts (e.g., Hebrew) are canonicalized to LTR by
mirroring $x$-coordinates ($x' = W - x$). The synthetic grid generator and the line-level reading-order
annotations for all 23 ALTO pages will be released publicly.

  
\begin{figure}[t]
  \centering
  \begin{minipage}{0.7\linewidth}
    \centering
    \begin{subfigure}{0.32\linewidth}
      \includegraphics[width=\linewidth]{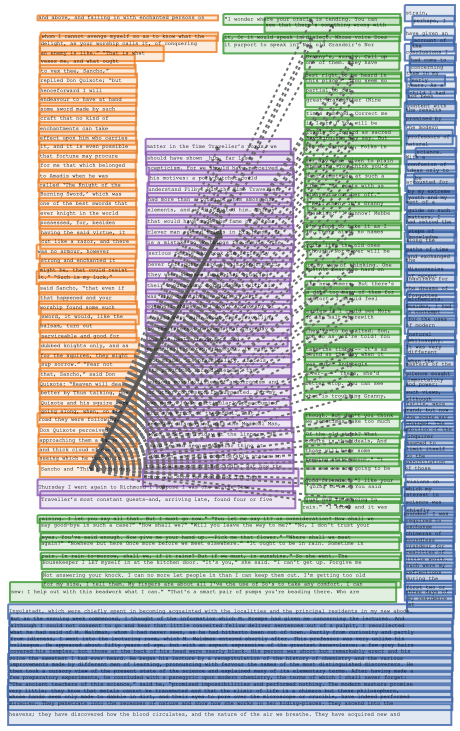}
      \caption{}
      \label{fig:synth_gt}
    \end{subfigure}\hfill
    \begin{subfigure}{0.32\linewidth}
      \includegraphics[width=\linewidth]{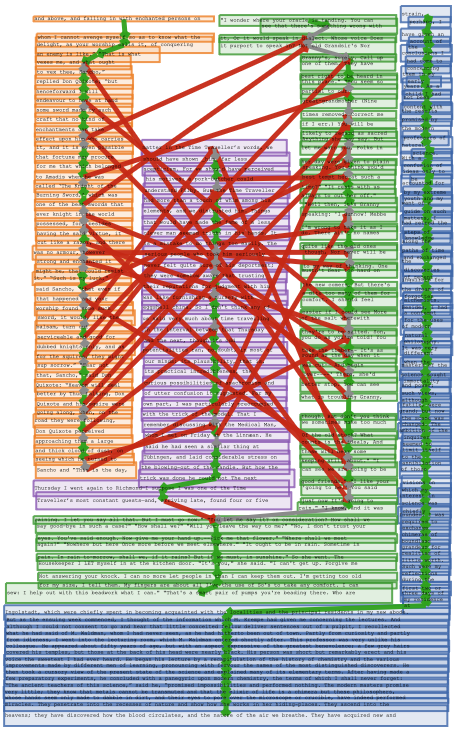}
      \caption{}
      \label{fig:synth_pred}
    \end{subfigure}\hfill
    \begin{subfigure}{0.32\linewidth}
      \includegraphics[width=\linewidth]{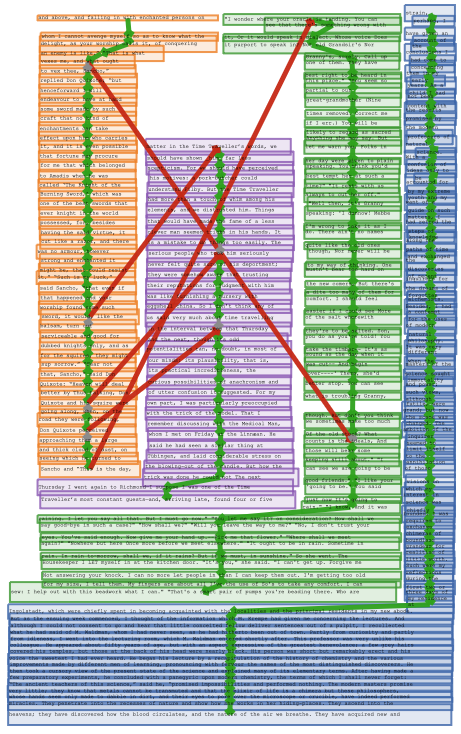}
      \caption{}
      \label{fig:synth_pred2}
    \end{subfigure}
  \end{minipage}
  \caption{Next-column stress test on an ALTO-based graph (5{,}783 candidate edges).
  (a) Candidate generator connects each line to all lines to the right of its right boundary.
  (b) Basic regret produces many spurious links.
  (c) Additive ensemble regret recovers the critical inter-stream continuation.}
  \label{fig:synth_qual_pair}
\end{figure}

\subsection{Metrics}
We report \emph{edge accuracy}: the fraction of non-terminal lines whose predicted successor matches
the ground-truth successor. The last line of each stream is excluded since it has no successor. Errors
are decomposed into \emph{same-stream skips} (predicted successor stays in the correct stream but skips
the immediate GT successor) and \emph{cross-stream links} (predicted successor belongs to a different
stream). Across the 23-page ALTO corpus, edge accuracy is computed over thousands of GT successor
edges and tens of thousands of candidate transitions (e.g., Fig.~\ref{fig:synth_qual_pair} alone
contributes 5{,}783 candidate edges). For OmniDocBench we additionally report the Normalized Edit
Distance metric defined in the OmniDocBench paper~\cite{omnidocbench} for direct comparability against
its published leaderboard.
\section{Results}
\label{sec:Results}

We compare our full method, the \emph{Weighted Additive Ensemble with Max-Regret inference},
against baselines spanning both the geometric and learning-based traditions. On synthetic grids we use
a \emph{Column Concatenation} baseline (nodes grouped by column, ordered top-to-bottom within
columns, concatenated left-to-right). On ALTO pages we benchmark against three literature baselines:
\emph{XY-Cut}~\cite{nagy1984icpr} as implemented in PaddleOCR PP-StructureV3
(\texttt{recursive\_xy\_cut}); \emph{LayoutReader} (LR)~\cite{wang2021layoutreader}, the layout-only
LayoutLMv3 model (\texttt{hantian/layoutreader}); and \emph{LayoutReader-T} (LR-T), the text+layout
seq2seq variant from \texttt{nielsr/layoutreader-readingbank}, fed the same OCR-line boxes and the same
line text that our pipeline uses. All four methods are evaluated on identical line-level inputs per page.

\subsection{Synthetic Grid Results}
On the $8\times8$ instance, performance rises from $50$ to $59.2 \pm 0.4$ correctly recovered successor
edges out of $61$ (geometry-only $\to$ ours). On $16\times16$, it rises from $230$ to $235.4 \pm 8.2$ out
of $253$ (Table~\ref{tab:baseline_comparison}). The smaller margin on the larger grid is expected: as
the synthetic layout grows, a larger fraction of the page becomes trivially recoverable by column-wise
ordering, leaving less headroom for semantic re-ranking.

\subsection{ALTO Results}
\label{sec:alto_results}

Table~\ref{tab:alto_aggregate} reports edge-accuracy averages over the 11 pages in our main ALTO test
corpus (10 historical source pages plus Sanity2Col), grouped by reading-order regime. The 10
vertical-flip and 2 horizontal-mirror variants are reserved for the robustness analysis in
Section~\ref{sec:mirror_inv}.

A clear three-regime picture emerges. On \emph{Manhattan-ceiling} pages (Asher, Pesach\_v2,
O\_IE104261952, Sanity2Col), XY-Cut recovers $100\%$ of GT edges; our method averages $96.0\%$, within
$11$ percentage points on each individual page. These pages confirm that the XY-Cut implementation is sound and expose the
only regime in which our method is not the top performer: the semantic ensemble occasionally re-orders
adjacent same-column lines when short line-level text fragments make local continuation probabilities
ambiguous.

On \emph{wrap-around / Glossa Ordinaria} pages (Lieberman, Y\_IE20854727, O\_IE102667218,
Y\_IE30700205), where the reading order does not respect simple whitespace partitions, the gap between
our method and XY-Cut is large: $94.8\%$ vs.\ $49.7\%$ on average. On Lieberman we recover
\textit{100.0\%} of GT edges vs.\ XY-Cut's \textit{39.6\%}; on Y\_IE20854727, \textit{98.9\%}
vs.\ \textit{40.5\%}. The failure mode of XY-Cut on these pages is the bridging-gloss mechanism
analyzed in Section~\ref{sec:analysis_lr}.

On \emph{mixed / near-Manhattan} pages (Berakhot2B, L\_IE104330339, Vatican Palatina), XY-Cut recovers
$91$\,--\,$99\%$ of GT edges and our method matches it within $\pm 4$ percentage points.
Across all 11 pages, the LayoutReader variants are uncompetitive: LR averages $22.5\%$ (range
$0$\,--\,$55$), LR-T $19.8\%$ (range $6$\,--\,$34$). The text channel does not close the gap; the
granularity mismatch with ReadingBank's word-level supervision is structural
(Section~\ref{sec:analysis_lr}).

\begin{table}[t]
\centering
\caption{Synthetic grid: geometry-only baseline vs.\ ours, mean $\pm$ std over 5 seeds.}
\label{tab:baseline_comparison}
\small
\setlength{\tabcolsep}{4pt}
\renewcommand{\arraystretch}{1.05}
\begin{tabular}{@{}lrrrrr@{}}
\toprule
& & \multicolumn{2}{c}{Column Concatenation} & \multicolumn{2}{c}{Ours} \\
\cmidrule(lr){3-4}\cmidrule(lr){5-6}
Dataset & GT edges & Correct & Acc.\ (\%) & Correct & Acc.\ (\%) \\
\midrule
$8\times8$   & 61  & 50  & 82.0 & $59.2 \pm 0.4$  & $97.0 \pm 0.7$ \\
$16\times16$ & 253 & 230 & 90.9 & $235.4 \pm 8.2$ & $93.0 \pm 3.2$ \\
\bottomrule
\end{tabular}
\end{table}

\begin{table}[t]
\centering
\caption{ALTO 11-page corpus: per-regime average edge accuracy (\%). Best per row in \textbf{bold}.}
\label{tab:alto_aggregate}
\small
\setlength{\tabcolsep}{5pt}
\renewcommand{\arraystretch}{1.05}
\begin{tabular}{@{}lrcccc@{}}
\toprule
Regime              & $n$ & Ours & XY-Cut & LR & LR-T \\
\midrule
Manhattan-ceiling   & 4 & $96.0$ & $\mathbf{100.0}$~\; & $27.9$ & $26.6$ \\
Mixed / near-Manhattan & 3 & $95.3$ & $\mathbf{96.1}$  & $15.7$ & $18.3$ \\
Wrap-around / Glossa & 4 & $\mathbf{94.8}$ & $49.7$  & $22.3$ & $14.1$ \\
\midrule
\textit{All 11}     & 11 & $\mathbf{95.4}$ & $80.5$ & $22.5$ & $19.8$ \\
\bottomrule
\end{tabular}
\end{table}

\subsection{Mirror-Invariance Robustness}
\label{sec:mirror_inv}

Reading-order methods should be \emph{geometry-equivariant}: mirroring or flipping a page should not
change which line follows which under the (mirrored) reading order. Methods trained on left-to-right
English may carry a directional prior that violates this. We evaluate this on two horizontal mirrors
(Y\_IE20854727, Y\_IE30700205) and 10 vertical flips of the source pages; ground-truth reading order is
re-derived under a column-major rule with edge-aligned clustering so the \emph{semantic} order is
preserved. The worst-case $|\Delta\text{Acc.}|$ across mirrors is $0.6$ percentage points for our method,
$2.1$ percentage points for XY-Cut, $8.7$ percentage points for LR, and $7.5$ percentage points for LR-T. LR-T \emph{is not} mirror-invariant,
consistent with its LTR-English training distribution and a concrete instance of the directional-prior limitation we discuss in Section~\ref{sec:analysis_lr}.

\subsection{Public-Benchmark Evaluation: OmniDocBench}
\label{sec:omnidocbench}

We further evaluate on a 140-page English multi-column subset of OmniDocBench~\cite{omnidocbench}
(92 double-column, 48 three-column; sources: 38 academic papers, 36 exam papers, 32 newspapers,
17 magazines, 15 books, 2 textbooks; 2{,}085 ground-truth successor edges). OmniDocBench annotates
reading order at the paragraph-block level rather than the OCR-line level that is the primary target
of this paper, so we evaluate at its native granularity (each annotated block is one node in our
graph). The graph formulation operates on (bbox, text) tuples regardless of unit size, so no
methodological change is required; the same ensemble weights from the synthetic-grid sweep are
transferred unchanged.

Table~\ref{tab:omnidocbench} reports per-source edge accuracy alongside the Normalized Edit Distance
metric used by the OmniDocBench paper itself~\cite{omnidocbench}. Two observations consistent with the
ALTO picture: (i) our method wins by a wide margin on the subsets where geometric partition struggles,
exam papers ($+54.7$ percentage points over XY-Cut) and academic literature ($+13$ percentage points), which are the multi-column
academic-style pages identified as challenging in prior reviews. (ii) XY-Cut wins on strongly
rectilinear layouts (newspapers, three-column books) where its projection profile finds clean gutters.
LayoutReader underperforms uniformly, even on subsets close to its ReadingBank training distribution
(academic literature: $37.2\%$), confirming the granularity-mismatch argument of
Section~\ref{sec:analysis_lr}: word-level supervision does not transfer to coarser units, whether OCR
lines or paragraph blocks.

The OmniDocBench leaderboard is populated entirely by end-to-end image-input
systems\footnote{MinerU, Mathpix, Marker; GOT-OCR, Nougat; GPT-4o,
Qwen2-VL-72B, InternVL2-76B~\cite{omnidocbench}. Best reported English NED
0.08 (MinerU), 0.12 (Qwen2-VL-72B), 0.13 (GPT-4o).}; ours is the first
reported result for the training-free (bbox + text) regime, with NED 0.24
on the multi-column subset and 0.10 on exam papers. We do not claim parity with
the image-input leaderboard (e.g., MinerU at 0.08): those systems solve a
different, full-image task, whereas our method uses only boxes and text.

\begin{table}[t]
\centering
\caption{OmniDocBench English multi-column subset (140 pages, 2{,}085 GT successor edges). Edge
accuracy (\%, higher is better) and Normalized Edit Distance (NED, lower is better) per data source. Edge accuracy is over the 140 pages all methods processed; NED is over the full 142-page subset.
Best per row in \textbf{bold}.}
\label{tab:omnidocbench}
\small
\setlength{\tabcolsep}{4pt}
\renewcommand{\arraystretch}{1.05}
\begin{tabular}{@{}lrcccccc@{}}
\toprule
                  & & \multicolumn{3}{c}{Edge accuracy (\%)} & \multicolumn{2}{c}{NED}  \\
\cmidrule(lr){3-5}\cmidrule(lr){6-7}
Source            & $n$ & Ours & XY-Cut & LR   & Ours  & XY-Cut \\
\midrule
academic\_lit.    & 38  & $\mathbf{93.4}$ & 80.4 & 37.2 & $\mathbf{0.14}$ & 0.27 \\
exam\_paper       & 36  & $\mathbf{96.9}$ & 42.2 & 10.9 & $\mathbf{0.10}$ & 0.62 \\
newspaper         & 32  & 74.2 & $\mathbf{96.2}$ &  8.9 & 0.51 & $\mathbf{0.07}$ \\
magazine          & 17  & $\mathbf{89.1}$ & 82.6 & 26.0 & $0.33$ & $\mathbf{0.30}$ \\
book              & 15  & 82.0 & $\mathbf{85.8}$ & 53.7 & 0.19 & $\mathbf{0.18}$ \\
textbook          & 2   & 85.7 & $\mathbf{100.0}$~ & 36.9 & 0.13 & $\mathbf{0.00}$ \\
\midrule
\textit{Macro}    & 140 & $\mathbf{88.0}$ & 75.3 & 24.6 & $\mathbf{0.24}$ & 0.30 \\
\bottomrule
\end{tabular}
\end{table}

\subsection{Ablations}

We sweep $(\alpha,\beta,\gamma)$ over 120 configurations on the $16\times16$ synthetic grid with 5
seeds and adopt $\alpha=1.0, \beta=0.2$, transferred unchanged to all ALTO and OmniDocBench
experiments; the additive ensemble with max-regret recovers $235.4 \pm 8.2$ of 253 GT edges
($93.0\%$). Performance is much more sensitive to $\alpha$ and $\beta$ than to $\gamma$ in the
tested range. The swept optimum in fact sets $\gamma=0$: the sentence-embedding term gives no
improvement, and when we force it active with a real similarity matrix its effect is at best neutral
and degrades accuracy monotonically as $\gamma$ grows (across the ALTO layouts edge accuracy never
rises and falls steadily at larger $\gamma$). The two language-model signals therefore carry the
method, the causal-LM term ($\alpha$) dominant and NSP ($\beta$) a consistent further gain, so the
deployed scorer omits the embedding. With the additive scores fixed at these weights, greedy edge
selection recovers only
$141.6 \pm 7.8$ of 253 GT edges ($56.0\%$), below the geometry-only baseline ($90.9\%$) and far below
max-regret ($93.0\%$). Replacing greedy with max-regret gains $+37.0$ percentage points on average. The error
decomposition in Table~\ref{tab:greedy_vs_regret} shows that greedy produces $27.8$ cross-stream and
$85.6$ same-stream errors per run; max-regret reduces both to $4.4$ and $9.2$ respectively.

\begin{table}[t]
\centering
\caption{Greedy vs.\ max-regret inference on the $16\times16$ grid using additive ensemble scores
($\alpha=1.0$, $\beta=0.2$), averaged over 5 seeds.}
\label{tab:greedy_vs_regret}
\small
\setlength{\tabcolsep}{5pt}
\renewcommand{\arraystretch}{1.05}
\begin{tabular}{@{}lrrrr@{}}
\toprule
Method & Correct & Acc.\ (\%) & Cross-stream & Same-stream \\
\midrule
Column Concatenation baseline & 230 & 90.9 & --- & --- \\
Greedy inference              & $141.6 \pm 7.8$ & 56.0 & 27.8 & 85.6 \\
Max-Regret inference          & $235.4 \pm 8.2$ & 93.0 & 4.4  & 9.2 \\
\bottomrule
\end{tabular}
\end{table}

\paragraph{Runtime.}
On a single NVIDIA A40, per-page inference averages $93.5$\,s (median $88$\,s; $10.5$\,--\,$290$\,s)
and peaks at $6.8$\,GB. Cost is dominated by per-edge semantic scoring, not the path search, and scales
with candidate-edge density, so dense pages with many short, adjacent lines are the most expensive.
\section{Analysis}
\label{sec:analysis_lr}

This section summarizes dominant failure mechanisms in dense multi-stream layouts and explains why
max-regret inference improves robustness and why the baselines we compare against fail in
structurally specific ways.

\subsection{Greedy vs.\ Max-Regret}
Constrained greedy enforces the same degree constraints as our main solver but commits edges by
absolute score (the globally best feasible edge at each step). In dense candidate graphs this is
vulnerable to \emph{premature commitment}: an early high-scoring but incorrect edge $(u,v)$ claims
target $v$, renders $v$'s true predecessor infeasible, and displaces other nodes onto weaker
alternatives, triggering cascades of cross-stream links rather than isolated local mistakes.
Max-regret retains the same feasibility constraints but reorders which decisions to commit first,
prioritizing high-opportunity-cost sources (nodes whose best and second-best outgoing candidates
differ greatly in score, Eq.~\eqref{eq:regret}) because delaying these ``must-take'' commitments
would let a competing node claim their best target.

\subsection{Baseline Failure Modes}

\paragraph{XY-cut needs a whitespace gutter that does not exist.}
Recursive XY-cut~\cite{nagy1984icpr} alternates horizontal and vertical projection-profile splits,
choosing a cut wherever there is a contiguous run of background larger than a configured gap
threshold. This requires a clean whitespace gutter fully separating the cells to be cut. On Glossa
Ordinaria and similar bridging-gloss pages, one or more boxes (a wide gloss line, a marginal
commentary that wraps across the gutter, or a full-width banner) intersect every candidate vertical
cut, so the projection profile never reaches background between would-be columns. The recursion then
either fails to split or splits on the wrong axis, after which every box in the misclassified region
receives an incorrect successor in the terminal scan order. Our method bypasses both failure modes because it never
decomposes the page into rectangles; a bridging row or shared whitespace band is just one more
candidate node.

\paragraph{LayoutReader is trained at the wrong granularity.}
LayoutReader~\cite{wang2021layoutreader} and the LR-T text+layout variant are trained on ReadingBank,
which provides reading-order supervision over \emph{words}. Our inputs are OCR text \emph{lines} or
paragraph blocks, each spanning many words and several pixels of vertical extent, so box geometry,
input sequence length, and box-to-token alignment all differ from training. We observe a consistent 6--34\% edge-accuracy band across all ALTO pages with
LR-T no better than LR despite being fed text. The mismatch is structural, not semantic; the text channel cannot make a word-level
seq2seq emit correct line or paragraph successors. The same mechanism explains the mirror-invariance failure in Section~\ref{sec:mirror_inv}: a model whose training distribution is LTR-English at word
granularity has no opportunity to learn reading order as a function of \emph{relative} layout.

\subsection{Semantic Failure Modes}
Three failure modes recur. \emph{Context-insensitive continuation}: fragments score high simply
because they are broadly probable, not because they truly follow $t_u$; the marginal subtraction in
Eq.~\eqref{eq:freqnorm} discounts globally likely fragments and emphasizes context-specific lift.
\emph{Cross-stream hallucination}: generic cues (quotation marks, dialogue verbs) appear across
multiple streams; combining NSP with CLM provides a complementary signal that
reduces these merges at high candidate density. \emph{Semantic teleportation}: permissive candidate
graphs allow semantically similar but spatially distant lines to link; candidate gating bounds this
in practice.  
\section{Limitations}
\label{sec:limitations}

\paragraph{Computational cost and score overlap.}
Our approach is training-free but not cost-free: scoring candidate edges requires running pretrained
models over many line pairs, with cost scaling in the number of candidates per node. Candidate gating
and score caching make this tractable in practice, but very dense candidate graphs (weak geometric
constraints or all-to-all candidates) remain expensive. Performance is also bounded by \emph{score
overlap}: short or semantically generic line fragments may produce near-identical scores for correct
and incorrect successors, making some errors irreducible without additional signals such as typography,
indentation, or learned layout priors. This is most acute for very short segments (marginalia,
page numbers, running heads) and degraded OCR, where every textual score weakens at once; fusing the
semantic scores with layout and visual cues, so the ensemble can fall back on geometry where language
is uninformative, is a natural extension.

\paragraph{Scope of evaluation.}
We deliberately isolate layout inference from OCR noise by using clean line boxes and controlled
English text. End-to-end behaviour under realistic OCR error rates therefore remains uncharacterized;
a CER/WER sweep on our populated ALTO pages is the most direct next-step robustness study. The
semantic stack (CLM + NSP) is also validated only on English: right-to-left
scripts, top-to-bottom and vertical writing systems, and historical language distributions with poor
tokenizer coverage are out of scope, though the graph formulation and inference rule are unchanged
under such substitutions. Our public-benchmark evaluation covers the multi-column English subset of
OmniDocBench (Section~\ref{sec:omnidocbench}) at its native paragraph granularity; extending this to
the full OmniDocBench corpus, to non-English subsets, and to line-level evaluations on other public
benchmarks is straightforward future work. Broader collections of annotated manuscripts and complex
business documents, such as the Tobacco-800 collection, would further stress-test robustness on diverse
real-world layouts. Our ensemble weights $(w_{\text{clm}}, w_{\text{nsp}})$ are likewise selected once
on the synthetic grid and transferred unchanged; per-layout-family tuning could improve individual
regimes at the cost of the cleanest transfer story.

\paragraph{Baseline-comparison gap.}
End-to-end document parsers such as Dots.OCR and PaddleOCR-VL operate at line granularity but bundle
detection, recognition, and reading order in a single pipeline, and do not expose their reading-order
module for evaluation on externally supplied OCR boxes. A clean apples-to-apples comparison on
identical inputs is therefore not currently possible; doing so would require either reimplementing
their reading-order head or running both systems on their own full-pipeline outputs, with the
attendant segmentation differences and partial-credit issues.
\section{Conclusion}
We presented a training-free framework for reading order inference in complex,
non-Manhattan layouts. OCR lines become nodes in a candidate transition graph;
edges are scored by a weighted additive ensemble of CLM likelihood and NSP.

On top of these local scores, a max-regret inference strategy outperforms greedy selection by
prioritizing high-opportunity-cost decisions and reducing catastrophic ``edge theft'' failures under
one-predecessor / one-successor constraints.

We evaluate across synthetic grids, 23 ALTO historical geometries, and a 140-page
OmniDocBench multi-column subset~\cite{omnidocbench}, against the canonical recursive
XY-cut (PaddleOCR PP-StructureV3) and two LayoutReader variants. Our method dominates on layouts where geometric partition is unreliable (Glossa, exam papers, academic), reaching 88.0\% accuracy (NED 0.24) on OmniDocBench versus XY-cut's 75.3\% (0.30) and
LayoutReader's 24.6\%, while structurally rectilinear layouts (newspapers,
clean Manhattan) remain in XY-cut's strength regime. Our pipeline is also mirror-invariant to within $\pm 0.6$ percentage points under horizontal and
vertical reflections, whereas LayoutReader-T is not.

Future work will focus on integrating lightweight visual cues, such as indentation and font styles, improving robustness against OCR noise, expanding support for multilingual historical texts, and extending OmniDocBench to the full corpus and other public benchmarks.

\subsubsection*{Acknowledgments.}
We thank the reviewers for their suggestions. This research was funded in part by the European Union as part of ``MiDRASH'' [\url{https://www.midrash.eu}] (ERC project no.\@ 101071829, with
principal investigators: Daniel St\"okl Ben Ezra, EPHE-PSL; Nachum Dershowitz, Tel Aviv University; Judith Olszowy-Schlanger, EPHE-PSL; and Avi Shmidman, Bar-Ilan University).
Views and opinions expressed are, however, those of the authors only and do not necessarily reflect those of the European Union or the European Research Council Executive Agency. Neither the European Union nor the granting authority can be held responsible for them.

\bibliographystyle{splncs04}
\bibliography{main}

\end{document}